%% file: main.tex
\documentclass[10pt,conference]{IEEEtran}
\IEEEoverridecommandlockouts
\usepackage{amsmath,amssymb,amsfonts}
\usepackage{algorithmic}
\usepackage{graphicx}
\usepackage{textcomp}
\usepackage{xcolor}
\def\BibTeX{{\rm B\kern-.05em{\sc i\kern-.025em b}\kern-.08em
    T\kern-.1667em\lower.7ex\hbox{E}\kern-.125emX}}
\usepackage{svg}
\usepackage[T1]{fontenc} 
\usepackage{microtype} 

\usepackage[english]{babel} 

\usepackage{hyperref} 
\usepackage{caption}
\usepackage{subcaption}
\usepackage{siunitx}
\usepackage{gensymb}
\usepackage{booktabs}
\usepackage{adjustbox}
\usepackage{tikz}
\usepackage{svg}
\graphicspath{ {images/} }
\usepackage[
    natbib=true,
    style=numeric,
    sorting=none,
    maxcitenames=5,
    maxbibnames=5
]{biblatex}

\usepackage[font=small,labelfont=bf]{caption}

\usepackage{glossaries}
\glsdisablehyper
\loadglsentries[main]{acronyms}

\begin{document}

\title{MARLIN: Soft Actor-Critic based Reinforcement Learning for Congestion Control in Real Networks
}

\author{
\IEEEauthorblockN{
Raffaele Galliera\textsuperscript{$\dag$, $\star$},
Alessandro Morelli\textsuperscript{$\dag$},
Roberto Fronteddu\textsuperscript{$\dag$},
Niranjan Suri\textsuperscript{$\dag$,$\star$,$\ddag$}}

\IEEEauthorblockA{\textit{\textsuperscript{$\dag$}Florida Institute for Human \& Machine Cognition (IHMC)}\\}

\IEEEauthorblockA{
    \textit{\textsuperscript{$\star$}Department of Intelligent Systems \& Robotics} -
    \textit{The University of West Florida (UWF)}\\
Pensacola, FL, USA
}

\IEEEauthorblockA{\textit{\textsuperscript{$\ddag$}US Army Research Laboratory (ARL)} \\
Adelphi, MD, USA \\
\{rgalliera, amorelli, rfronteddu, nsuri\}@ihmc.org}
}

\newcommand\copyrighttext{%
  \footnotesize \textcopyright 2023 IEEE. Personal use of this material is permitted.
  Permission from IEEE must be obtained for all other uses, in any current or future
  media, including reprinting/republishing this material for advertising or promotional
  purposes, creating new collective works, for resale or redistribution to servers or
  lists, or reuse of any copyrighted component of this work in other works.}
\newcommand\copyrightnotice{%
\begin{tikzpicture}[remember picture,overlay]
\node[anchor=south,yshift=10pt] at (current page.south) {\fbox{\parbox{\dimexpr\textwidth-\fboxsep-\fboxrule\relax}{\copyrighttext}}};
\end{tikzpicture}%
}

\maketitle
\copyrightnotice
\begin{abstract}
Fast and efficient transport protocols are the foundation of an increasingly distributed world. The burden of continuously delivering improved communication performance to support next-generation applications and services, combined with the increasing heterogeneity of systems and network technologies, has promoted the design of Congestion Control (CC) algorithms that perform well under specific environments. The challenge of designing a generic CC algorithm that can adapt to a broad range of scenarios is still an open research question. To tackle this challenge, we propose to apply a novel Reinforcement Learning (RL) approach. Our solution, MARLIN, uses the Soft Actor-Critic algorithm to maximize both entropy and return and models the learning process as an infinite-horizon task. We trained MARLIN on a real network with varying background traffic patterns to overcome the sim-to-real mismatch that researchers have encountered when applying RL to CC. We evaluated our solution on the task of file transfer and compared it to TCP Cubic. While further research is required, results have shown that MARLIN can achieve comparable results to TCP with little hyperparameter tuning, in a task significantly different from its training setting. Therefore, we believe that our work represents a promising first step towards building CC algorithms based on the maximum entropy RL framework.
\end{abstract}


\section{Introduction}
    \input{sections/introduction}

\section{Designing a Reinforcement Learning agent for Congestion Control}
   \input{sections/agent_design}
    \begin{table}[t]
            \centering
            \input{tables/hyperparameters}
            \caption{Hyperparameters used in MARLIN.}
            \label{tab:hyperparameters}
    \end{table}
        
\section{The Mockets Transport Protocol}
    \label{sec:mockets}
    \input{sections/mockets_communication_protocol}
    
\section{Experimental Results}
    \label{sec:results}
    \input{sections/experimental_results}

\section{Related Work}
    \input{sections/related_work}

\section{Conclusion}
    \input{sections/conclusion}
    
    \begin{figure}[h]
        \centering
        \begin{subfigure}[b]{\linewidth}
        \includegraphics[width=\linewidth]{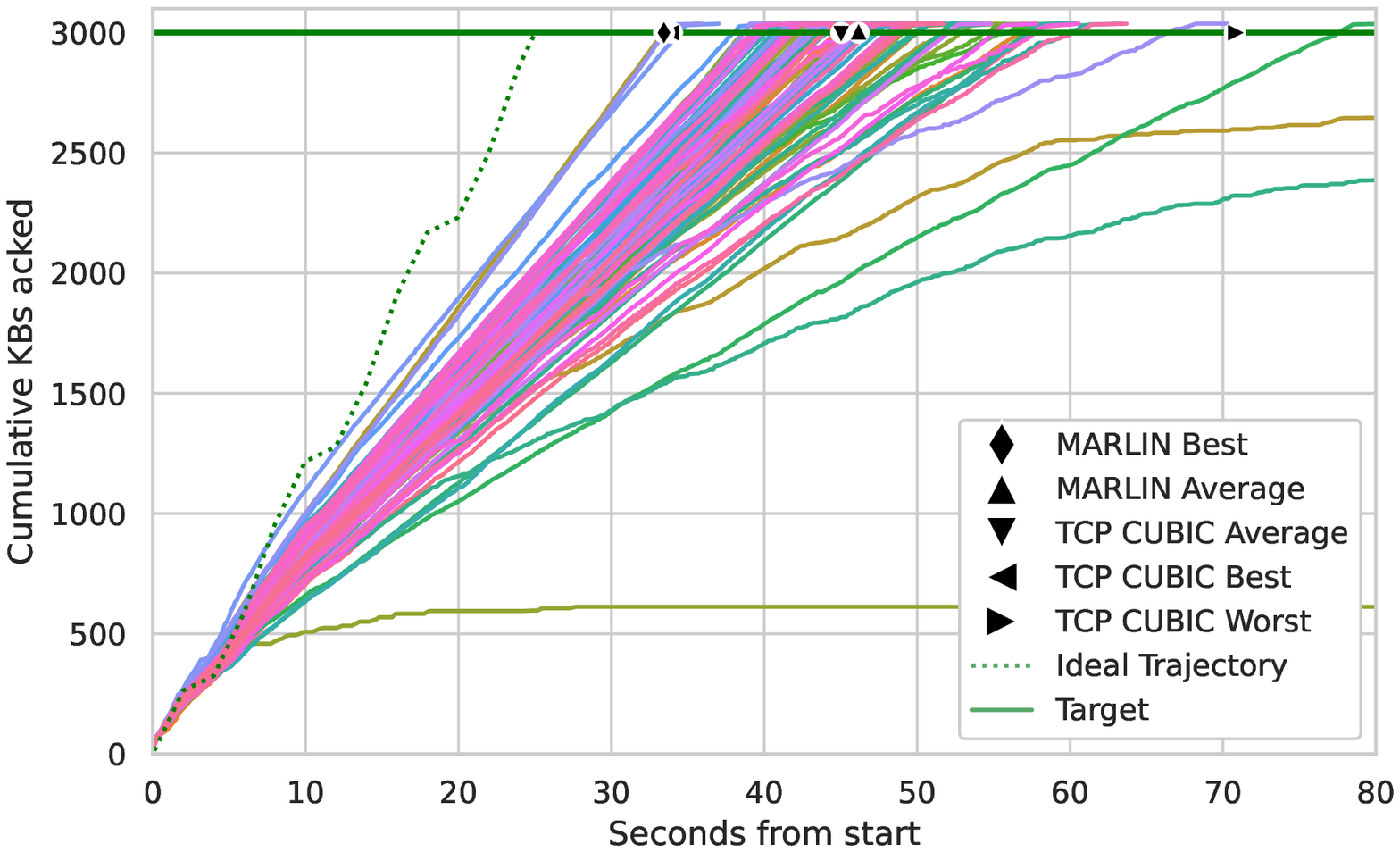}
        \smallskip
        \subcaption{Agent trained on a single traffic pattern.}
        \label{marlin_a}
        \end{subfigure}
        
        \begin{subfigure}[b]{\linewidth}
        \includegraphics[width=\linewidth]{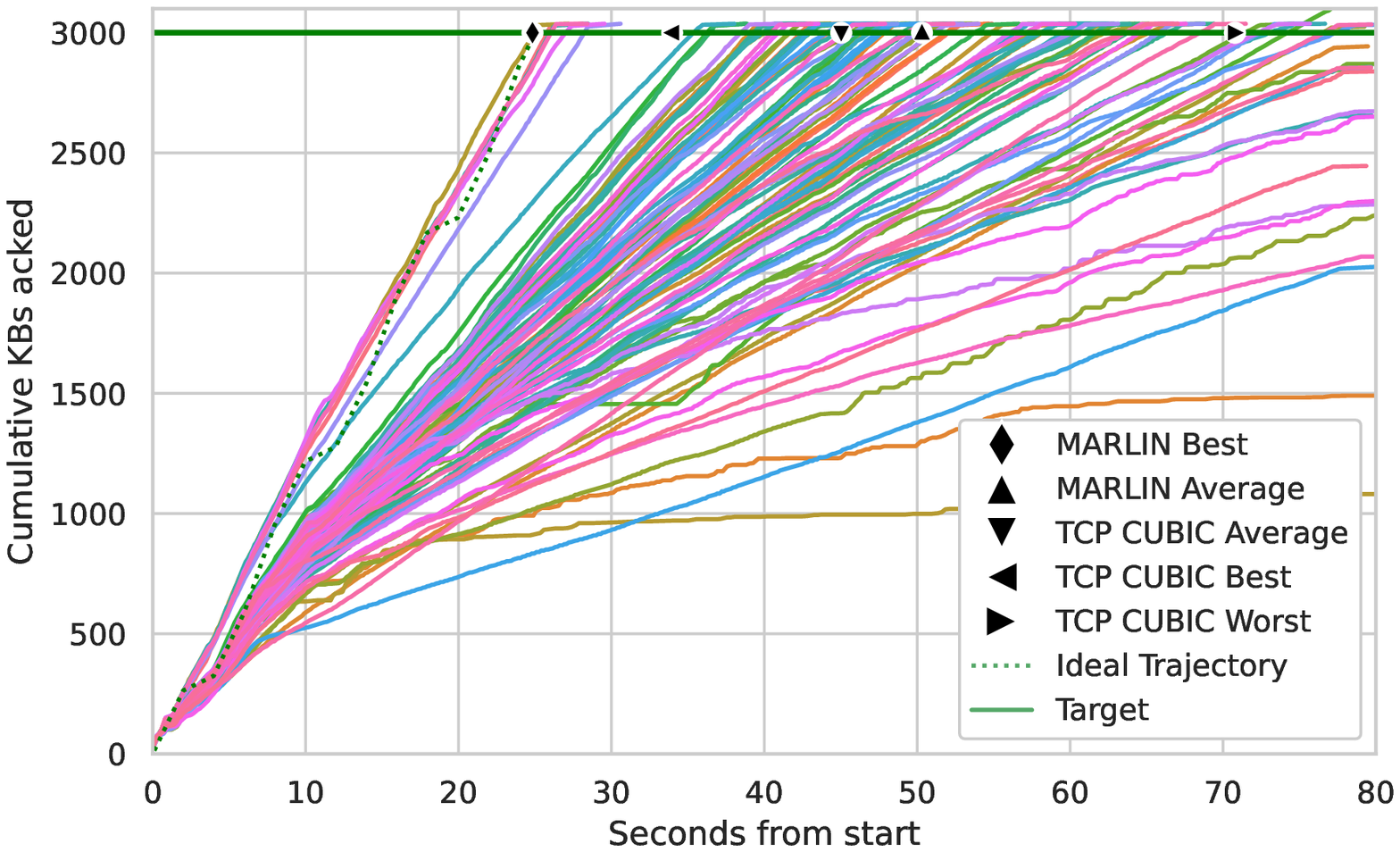}
        \smallskip
        \subcaption{Agent trained on a single traffic pattern with RTT penalties.}
        \label{marlin_b}
        \end{subfigure}
        
        \begin{subfigure}[b]{\linewidth}
        \includegraphics[width=\linewidth]{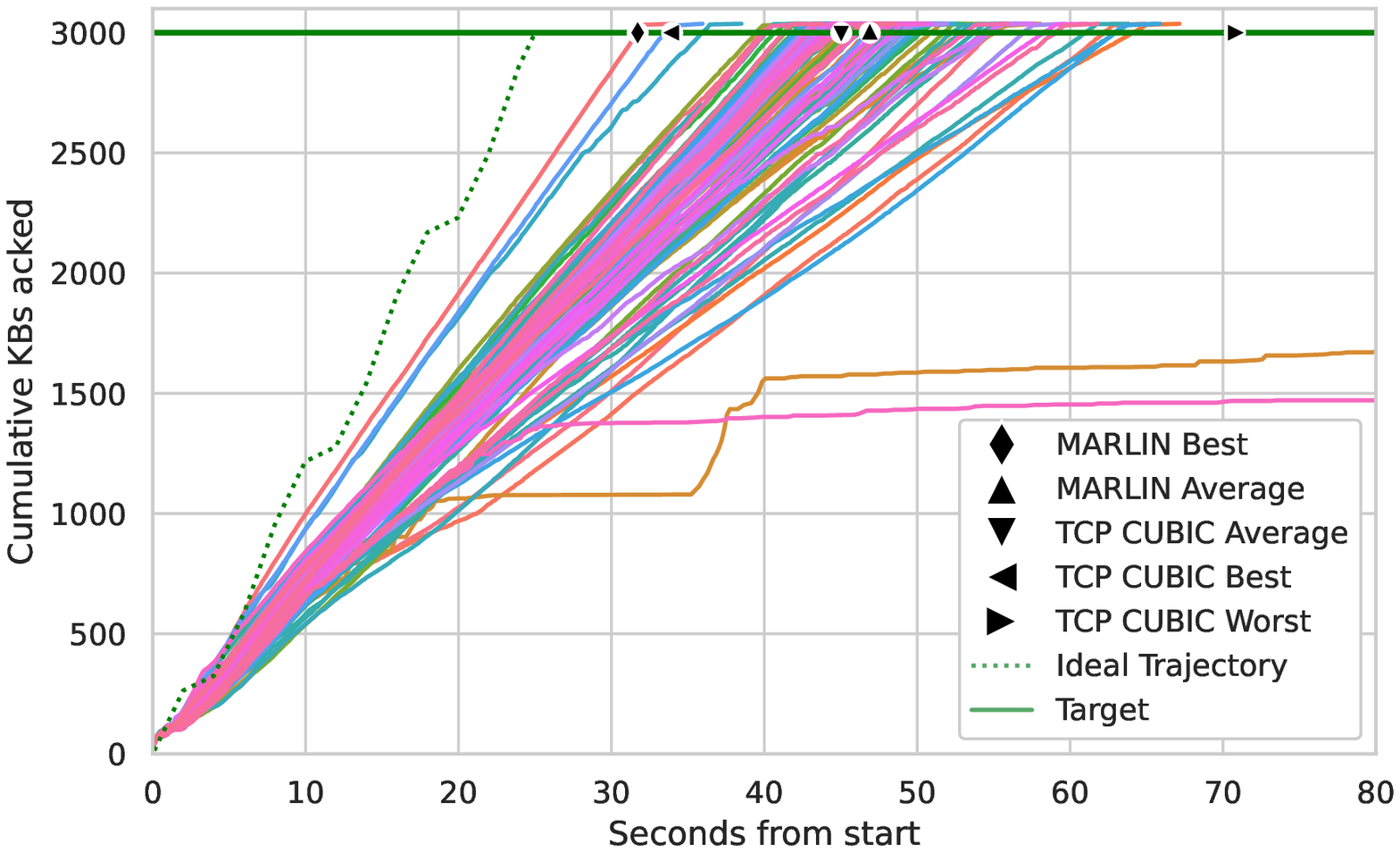}
        \subcaption{Agent trained on every permutation of the traffic flows.}
        \label{marlin_c}
        \end{subfigure}

        \caption{Evaluation of each model on 100 testing experiments.}
        \label{fig:trajectories}
    \end{figure}
\newpage
\printbibliography

\end{document}

%% file: acronyms.tex
\newacronym{cwnd}{CWND}{Congestion Window}
\newacronym{rl}{RL}{Reinfocement Learning}
\newacronym{rtt}{RTT}{Round-Trip Time}
\newacronym{srtt}{SRTT}{Smoothed Round-Trip Time}
\newacronym{bdp}{BDP}{Bandwidth Delay Product}
\newacronym{mdp}{MDP}{Markov Decision Process}
\newacronym{sb3}{SB3}{Stable-Baselines3}
\newacronym{ack}{ACK}{ACKnowledge}
\newacronym{rpc}{RPC}{Remote Procedure Call}
\newacronym{sac}{SAC}{Soft Actor-Critic}
\newacronym{ema}{EMA}{Exponential Moving Average}
\newacronym{ml}{ML}{Machine Learning}
\newacronym{ai}{AI}{Artificial Intelligence}
\newacronym{cc}{CC}{Congestion Control}
\newacronym{tcp}{TCP}{Transmission Control Protocol}
\newacronym{rl3}{RL-Zoo3}{RL Baselines3 Zoo}
\newacronym{mgen}{MGEN}{Multi-Generator Network Test Tool}
\newacronym{ap}{AP}{Access Point}
\newacronym{cots}{COS}{Commercial Off-The-Shelf}
\newacronym{rpi}{RPi4B}{Raspberry Pi 4 Model B}
\newacronym{qos}{QoS}{Quality Of Service}
\newacronym{peb}{PEB}{Partial Episode Bootstrapping}
\newacronym{scp}{SCP}{Secure Copy Protocol}
\newacronym{netem}{TC-NETEM}{TC NetEm - Traffic Control Network Emulator}
\newacronym{asic}{ASIC}{Application-Specific Integrated Circuit}
\newacronym{sgd}{SGD}{Stochastic Gradient Descent}
\newacronym{marl}{MARL}{Multi-Agent Reinforcement Learning}
\newacronym{nic}{NIC}{Network Interface Card}

%% file: sections/introduction.tex
Network communications are the backbone of an ever-increasingly distributed digital world. Each day, technologies such as software, platform, and infrastructure as a service over Cloud/Edge/Fog computing systems, the Internet of Things, 5G networks, space communications and networks, peer-to-peer networks, blockchain, ad-hoc and mesh networks enable countless organizations and people to run their business and perform tasks that have become part of our daily routine with just a tap on a screen. The different characteristics of those technologies, combined with the requirements of very diverse services built on top of them, present unique challenges for network protocol designers and researchers.

Transport protocols that can achieve high network resource utilization while maintaining congestion, i.e., end-to-end queueing delays, low in presence of ever-changing network conditions and high churn of connections are the key enabling technology that underlies modern distributed systems. Within transport protocol design, this dual goal (high channel utilization and low congestion) is the objective of \gls{cc} algorithms. The critical impact of \gls{cc} on the performance of distributed applications and services has led researchers and engineers to design many variants, each one with different target scenarios and trade-offs in terms of aggressiveness, responsiveness to loss and congestion, fairness, and friendliness \cite{lorincz2021, ma2017, widmer2001}. However, the challenge of designing a generic \gls{cc} algorithm that can provide near-optimal performance in a broad range of network and traffic scenarios is still open.

Recent advances in computational capacity, made possible by novel CPU, GPU, and \gls{asic} architectures, along with the availability of low-cost, yet powerful, versions of such hardware accelerators, have led to an explosion of successful applications of \gls{ml} technology in several domains, coming to meet the requirements of resource constrained environments such as the edge~\cite{GALLIERA2022239}. This has sprung renewed research interest towards the development of \gls{ml} models, \gls{rl} agents, algorithms, and applications that can face the unique challenges of the real-world~\cite{anymal, openAI_dexterous, sac_applications, delta21}.

The difficulty of designing an efficient generic \gls{cc} algorithm, in contrast with the relative ease of collecting data on the communication performance and the small size of the action space, which typically focuses solely on adjusting the size of the \gls{cwnd}, has prompt researchers to investigate the use of \gls{ml} for \gls{cc} optimization~\cite{Wei2021, Jiang2021}. While many studies have already shown promising results, they still underperform when compared to existing TCP \gls{cc} algorithms in many scenarios. Reasons may include: difficulties to train \gls{rl} agents and/or \gls{ml} models using real networks and hardware; low fidelity of simulation and emulation environments; sub-optimal environment representation (state), action space, and/or reward function design; and delayed action effects on the state due to communications delay. Additional research is needed to address these problems.

This paper introduces our initial work on MARLIN, a \gls{rl} agent for \gls{cc}. Our approach is based on \gls{sac}~\cite{sac}, an off-policy, entropy-regularized \gls{rl} algorithm, that has seen successful application in numerous real-world problems, especially robotics~\cite{sac_applications}. We describe the design and architecture of MARLIN and discuss how we trained the agent, the hyperparameters used, and the main assumptions and choices we made during its design phase, comparing our work with other \gls{rl}-based approaches to \gls{cc}. Furthermore, we discuss the future directions of this research, of which this paper is just the first step. Finally, we present preliminary experimental results obtained in a real networking scenario, where we investigate the current performance of MARLIN and compare it to the \gls{tcp} in handling a file transfer task.

%% file: sections/agent_design.tex

MARLIN is a \gls{rl} agent for \gls{cc}, trained in a real network by using continuous actions to update the \gls{cwnd}. For the purpose of this research, we did not consider fairness towards competing MARLIN flows and/or other protocols as one of the optimization objectives of MARLIN, which we leave as a future research question.

\begin{figure}[h]
        \centering
        \includegraphics[width=\columnwidth]{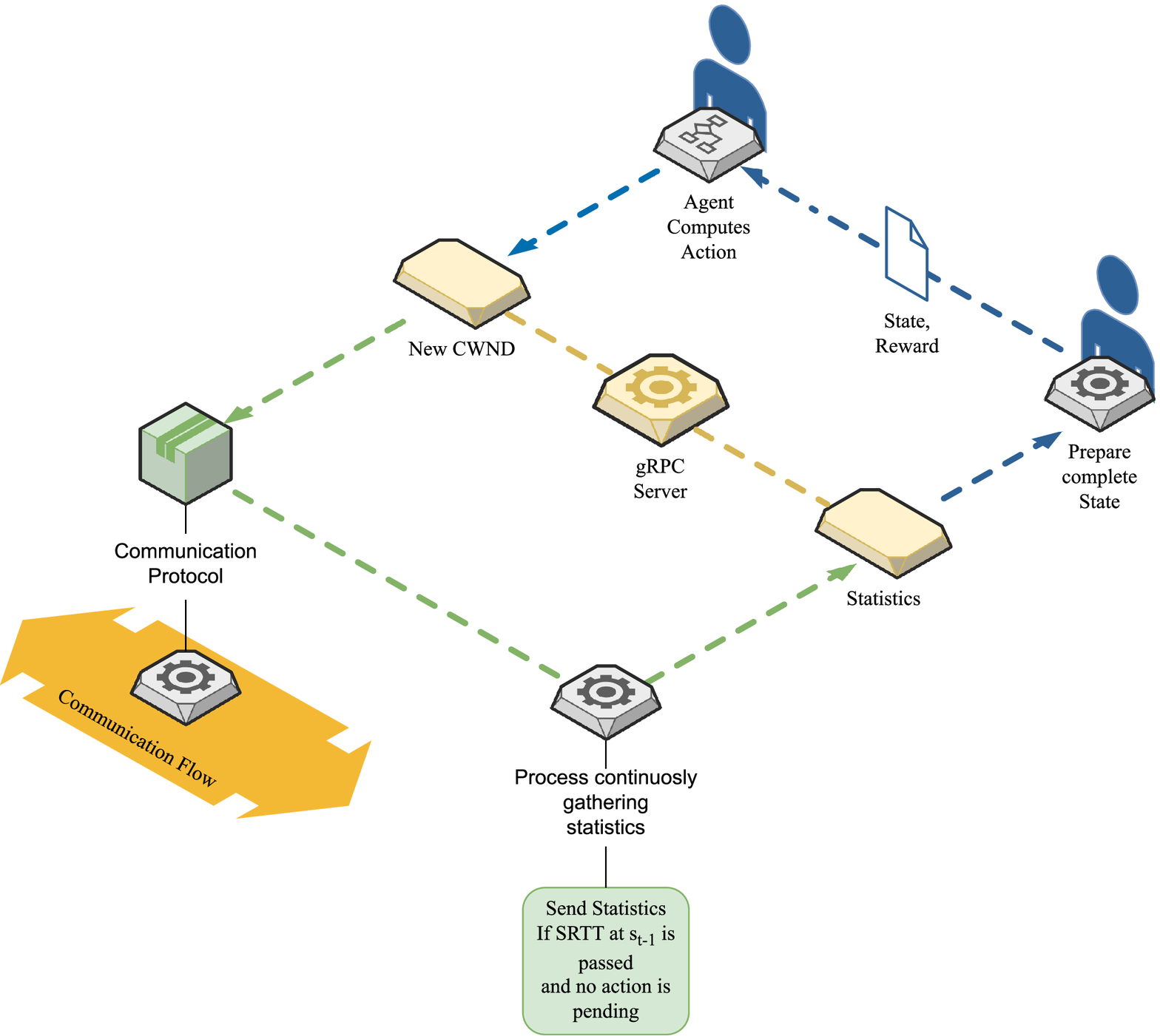}
        \caption{Agent-Protocol interface. Communication protocol and agent communicate through a gRPC server.}
        \label{fig:architecture}
\end{figure}

Within MARLIN, learning is based on \gls{sac}~\cite{sac}, which trains a stochastic, off-policy, entropy-regularized agent. The agent interfaces with a custom transport protocol, presented in Section~\ref{sec:mockets}, via \glspl{rpc} in a non-blocking, bi-directional fashion, as shown in Figure~\ref{fig:architecture}. The frequency of action taking depends on the latest \gls{srtt} estimated for the path. The reward function is shaped as a strictly negative reward to intrinsically encode the principle that \textit{every step taken to transfer the information is a step too much}.

Although it is generally desirable that the agent completes its tasks as quickly as possible, the \gls{rl} setting proposed here does not aim for terminal states. It is inconvenient for the agent to overfit its trajectories on a fixed amount of bytes to transfer, or a fixed episode length, such that it will try to find a terminal state at any cost in a certain number of timesteps. Instead, we aim at designing \gls{rl} agents able to work on "crowded" links and achieve high channel utilization independently from the volume of data transferred during training. For such reason, training is based on infinite-horizon tasks, never encountering a terminal state and following the \gls{peb} principles for infinite-horizon training presented in \cite{time_limit}.

\begin{figure*}
\centering
\includegraphics[width=\textwidth]{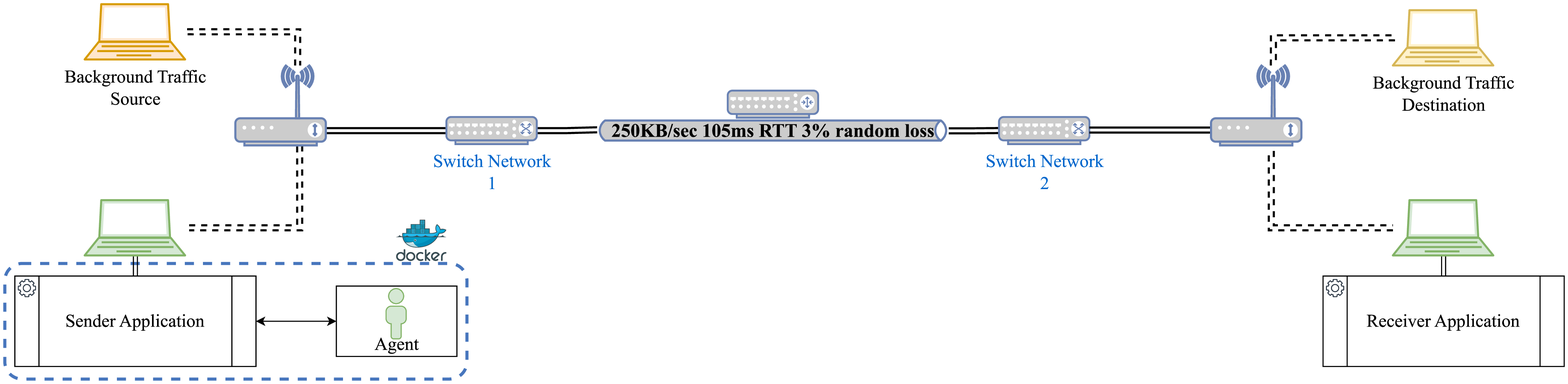}
\caption{A dumbbell network is used for training and testing. Background and agent traffic are transmitted from one sub-network to the other over the bottleneck link. Senders and receivers are all connected to a WiFi Access Point.}
\label{fig:network}
\end{figure*}

\subsection{Congestion Control as Markov Decision Process}
    Traditional \gls{cc} algorithms implemented within transport protocols such as \gls{tcp} take actions, e.g., updating their \gls{cwnd}, in response to changes in the environment they sense with the goal of achieving maximum throughput and minimum congestion. Changes may include, for example, packet loss, duplicate packets, or variations in the measured \gls{rtt}. Such sequential decision making setting can be naturally posed as a \gls{mdp}~\cite{puterman1994}.
    
    The \gls{mdp} framework formalizes with $\langle S, A, p, R \rangle$ the sequential agent-environment interaction as the exchange of three different signals at every time step $t$ of the process. The \gls{rl} \textit{"learning from interaction"} fundamentals~\cite{suttonBarto} stem from this abstraction: the \textit{action} $a_t \in A(s_t)$ taken by the agent at a certain step $t$ after sensing the \textit{state} $s_t \in S$ leads to a consequent response of the environment as a \textit{reward} signal $r_{t+1} \in R \subset \mathbb{R}$ and transitions to a new state $s_{t+1}$. This process follows the environment dynamics described by the probability function $p(s_{t+1}, r_{t+1}|s_t, a_t): S \times R \times S \times A \longrightarrow [0, 1]$. Such formulation of the transition dynamics also defines the \textit{Markov property}, which requires the state to include all the relevant information about past interactions that would impact the future, as each possible values of $r_t \in R$ and $s_t \in S$ depend solely on the previous state and action $s_{t-1}, a_{t-1}$.
    
    Communication networks are a subtle environment for a \gls{rl} agent. Packets in modern networks are often transmitted a few milliseconds apart, sometimes microseconds. At the same time, running \gls{sgd} to update the model during training, or even simply computing a new action, can take up a significantly greater amount of time. As a consequence, the environment might have already changed significantly by the time the action is taken, potentially making it obsolete or, even, wrong. This has been identified as one of the main reasons why it is hard to maintain the performance achieved in a simulated environment after moving to a real-world deployment \cite{zhang2020}.
    
    Furthermore, actions do not instantly affect the agent's perception, as the effects of changing the \gls{cwnd} require some time before they have an impact on the network and that impact is propagated back to the source. As new packets are transmitted at a different rate after an action has been taken, they first need to reach their destination, then the receiver needs to generate and transmit a new \gls{ack} message for those packets, and finally that \gls{ack} needs to arrive back at the source before any information about the impact of the action taken can be inferred. These events will take, at the very least, one \gls{rtt} to complete. This raises a fundamental question for anybody who wants to apply \gls{rl} to the problem of \gls{cc}: \textit{when should the next action be taken?}
    While other similar approaches, like \cite{Sivakumar2019, NEURIPS2019_f69e505b}, gather statistics during a fixed amount of time to compute the state and take the next action, MARLIN implements a different heuristic algorithm. Every time the transport protocol makes new network statistics available to MARLIN, if at least the last reported \gls{srtt} has elapsed since the time the last action was taken, data collected up to that time is refined, the reward $r_{t}$ with regards to action $a_{t-1}$ is assigned, a new state $s_t$ is fed to the agent, and a new action $a_{t}$ is taken. Waiting for at least one \gls{srtt} between actions prevents that a new action is taken before the impact of the previous one can be sensed by the agent.
     
    \begin{table*}[h]
    \centering
    \small
    \begin{adjustbox}{}
        \input{tables/state}
    \end{adjustbox}
    \caption{Features composing the state of the environment. The horizon of the observation is also augmented with the previous 10 observations.}
    \label{tab:state}
    \end{table*}
    
    \subsection{Training Environment Setup}
    \label{subsec:training_env_setup}
    A key design point of MARLIN is to be trained and evaluated on real networks.
    Utilizing real components allowed us to understand the challenges that the \gls{cc} domain brings to \gls{rl}, avoiding the risk of failing a "sim-to-real" transposition, which could be caused, by inaccurate emulated/simulated patterns and weak assumptions, at the cost of a slower training time. Relying on real hardware and protocol implementation intrinsically diversifies training experience thanks to the stochasticity of the environment and it prevents the agent from overfitting on artificial patterns that would not be found in the real world.
    
    To this end, we built the network shown in Fig. \ref{fig:network}, which comprises two sources of traffic and two receivers, two WiFi \gls{ap}, two network switches, and one router connected in a dumbbell topology. Traffic sources are connected to the WiFi \gls{ap} located at one end of the dumbbell, while traffic receivers are connected to the \gls{ap} at the other end. The "Background Traffic Source", on the left, is responsible for generating background and send it to the "Background Traffic Receiver" node, on the right. The "Sender Application" transmits traffic to the "Receiver Application" via a custom transport protocol implementation that uses MARLIN as its \gls{cc} algorithm.
    
    The need for implementation flexibility led us to integrate MARLIN with Mockets, a protocol originally designed for communication environments characterized by limited bandwidth, typically found in tactical and wireless sensor networks. Further details regarding Mockets are presented in Section~\ref{sec:mockets}. To avoid running into situations where MARLIN could underperform due to the protocol implementation, we shaped the network scenario around the typical Mockets use case for this first iteration. 
    We set a Smart Queue policy on the router to limit the maximum amount of traffic flowing between the two subnetworks to 250 KB/s; we also used \gls{netem} on the "Sender Application" node to introduce 100 ms of latency and 3\% random packet loss that only affect the traffic generated by the application. Note that the router's manual reports that it effectively limits the actual rate to 95\% of the specified value when Smart Queue is enabled~\cite{edgeos_user_guide}.
    
    The background traffic patterns generated can be divided into elephant flows, i.e., long-lived data transfers that represent a large percentage of the total traffic (imagine large file transfers or video streaming), and mice flows, i.e., short-lived data transfers at low throughput (e-mail or \gls{rpc} calls). Roughly 87\% of the background traffic in our testbed is made by elephant flows, consisting of four different flows that alternate each other over the link every two seconds, and two mice flows, which continuously generate very short-lived (in the order of milliseconds) traffic bursts with intervals that follow a Poisson distribution.
    
    The Background Traffic Source generates traffic using \gls{mgen}~\cite{mgen}. Elephant flows consist of two UDP communications transferring data at 100 KB/s, one UDP communication transferring data at 50 KB/s, and one TCP connection producing 200 KB/s of traffic. One TCP and one UDP mice flow introduce an extra 17 KB of traffic in average over the link each second. Each elephant flow is assigned to a different temporal slot of a 2 second duration and they repeat every 8 seconds. Mice flows start with the first elephant flow and continue until traffic generation is stopped.


    
    

\subsection{Agent Design}
    The interface between MARLIN and the transport protocol, shown in Figure~\ref{fig:architecture}, is implemented using \textit{gRPC}. The gRPC middleware sitting between the transport protocol and MARLIN takes care of delivering observations and reward pairs to the agent and actions back to the protocol.
    
    At a given step $t$, the agent receives network statistics from the transport protocol via the gRPC middleware. Statistics are then processed and stacked with the previous 10 observations to form the state $s_t$. The agent can take actions that will increase, maintain constant, or decrease the transport protocol \gls{cwnd} by a chosen factor. Ideally, the agent should learn to maximize the volume of data transmitted in the minimum amount of time, while being watchful of growing queueing delays, which would manifest with an increase in the measured \gls{rtt}.
    
    MARLIN is implemented on top of the \gls{rl3}~\cite{rl-zoo3} framework, which follows best practices for using \gls{sb3}~\cite{stable-baselines3}, a PyTorch~\cite{pytorch}-based library that implements state-of-the-art \gls{rl} algorithms following the OpenAI gym interface~\cite{openAI_gym}.
    
    \subsubsection{State}
        Table~\ref{tab:state} describes the state encoded in MARLIN. 14 features are gathered during the time frame that follows the action taken at step ${t-1}$. MARLIN then augments the state space with 7 statistics, i.e., last, mean, standard deviation, minimum, maximum, \gls{ema}, and difference from the previous state $s_{t-1}$, that are computed for each of the 14 features.
        The previous $N$ states are also stacked together to form a history of the previous observations, attempting to adhere to the Markov property. The final state served to the agent will then have $N \times |Features| \times |Stats|$ features. This totals up to 980 different features.
        
        $N$ is considered a hyperparameter of the problem; table~\ref{tab:hyperparameters} has a complete list of all hyperparameters used for MARLIN. For the choice of $N$, we followed the empirical considerations made in~\cite{pmlr-v97-jay19a}, and chose 10 as the length of the history after some preliminary trials and evaluations. During the training process, observations are processed and normalized through a moving average by using the \textit{VecNormalize} environment wrapper present in \gls{sb3}.

    \subsubsection{Actions}
        MARLIN takes continuous actions contained in the range $[-1, 1]$, which represent the percentage gain of the \gls{cwnd} size. For example, if the action chosen by the agent is going to be $1$, the \gls{cwnd} is doubled; a value of $0$ means no change in the \gls{cwnd}; $-0.5$ reduces the window size by 50\%. The initial \gls{cwnd} size is set to 4KB at the beginning of the episode, as per the transport protocol implementation default.
        
        For the purpose of this study, we capped the \gls{cwnd} size to 50 KB, a value that, if reached, would yield double the throughput that the network can accommodate in the setup we used for training and evaluation. This choice only impacts the very first phases of training, when the agent takes random steps to prime the model. After this phase, which in MARLIN is set to last 10K steps, our experiments have shown that the agent correctly never fills the \gls{cwnd} to 50 KB.

    
    \subsubsection{Reward}
        
        
        
        

        We designed a reward function that gives higher rewards to the agent the closer it gets to fully utilizing the available bandwidth:
        
        \begin{equation}
        \label{eq:reward_1}
            r_t = -\frac{target_t}{target_t + acked\_kilobytes_t^{cumulative}}
        \end{equation}
        
        where $target_t$ represents the amount of bytes the agent should have delivered up to step $t$ since the beginning of the episode in order to fully utilize the link and $acked\_kilobytes_t^{cumulative}$ represents the number of kilobytes there were acknowledged by the receiver until step $t$.
        
        A strictly negative reward function promotes the agent to accumulate the smallest amount of penalties. The penalty received is much smaller the closer the agent it is, at each step, to having utilized the link to the best of its possibilities.
        
        The careful reader might notice that such rewarding system encourages the agent to accumulate acked bytes regardless explicit impact on the \gls{rtt}, falling into the risk of privileging actions that could produce more acked bytes in the immediate future, with the drawback of causing undesired queuing delays.
        To prevent such risk, we consider a second formulation of the reward function that introduces a \gls{rtt}-based penalty coefficient:
        
        \begin{equation}
        \label{eq:reward_2}
            r_t = -\frac{target_t}{target_t + acked\_kilobytes_t^{cumulative}*(1 - penalties)}
        \end{equation}
        
        The term $penalties$ depends on the difference between the current \gls{rtt} and the minimum \gls{ema} \gls{rtt} and is defined as follows:
        \begin{equation}
        \label{eq:penalties}
            penalties = \begin{cases} 
            
            \alpha \frac{rtt_{diff}}{rtt_{min}^{ema}}, & \mbox{if } \frac{rtt_{diff}}{rtt_{min}^{ema}} < 1 
    
            \\ 0.99, & \mbox{otherwise}
            \end{cases} 
        \end{equation}
        
        In Eq. \ref{eq:penalties}, $\alpha$ depends on the magnitude of the difference between $rtt_{diff}$ and $rtt_{min}^{ema}$ and it is defined as follows:
        
        \begin{equation}
            \label{eq:alpha}
            \alpha = \begin{cases} 
            
            1, & \mbox{if } |\frac{rtt_{diff}}{rtt_{min}^{ema}}| > 0.6 
            
            \\ 0.5, & \mbox{if } 0.1 < |\frac{rtt_{diff}}{rtt_{min}^{ema}}| \le 0.6
            
            \\ 0.3, & \mbox{if } 0.05 < |\frac{rtt_{diff}}{rtt_{min}^{ema}}| \le 0.1
            \\ 0.1, & \mbox{otherwise}
            \end{cases} 
        \end{equation}
        
        It is worth noticing that, in case $rtt_{diff} < 0$, the $penalties$ term becomes negative, thus rewarding the agent when \gls{rtt} improvements are detected.
    
    \subsubsection{RL Algorithm}
         MARLIN adopts \gls{sac}~\cite{sac}, an off-policy actor-critic algorithm based on the maximum entropy \gls{rl} framework. \gls{sac} augments the maximum reward objective, foundation of \gls{rl} settings, with an entropy maximization term. Such term acts as a trade-off between exploration and exploitation, so that the agent aims to maximize its return while also acting as random as possible. In circumstances where multiple actions seem equally attractive, i.e. in the case of equal or close Q-Values, the learned policy is encouraged to assign equal probability mass to those actions.
    
         In practice, the effects of the entropy, or temperature, term prompt the agent to discard unsuitable trajectories in favor of more promising ones, as well as to improve the learning speed. The entropy term can be either fixed or optimized/learned as further steps are taken. However, the optimal entropy coefficient varies depending on a series of factors, such as the nature of the task or even the current policy. As a consequence, it is usually considered good practice to avoid fixed values, preferring instead to update the term at the same time actor, critic, and the target networks are optimized~\cite{sac_applications}.
    

\begin{figure*}
\centering
\includegraphics[width=0.8\textwidth]{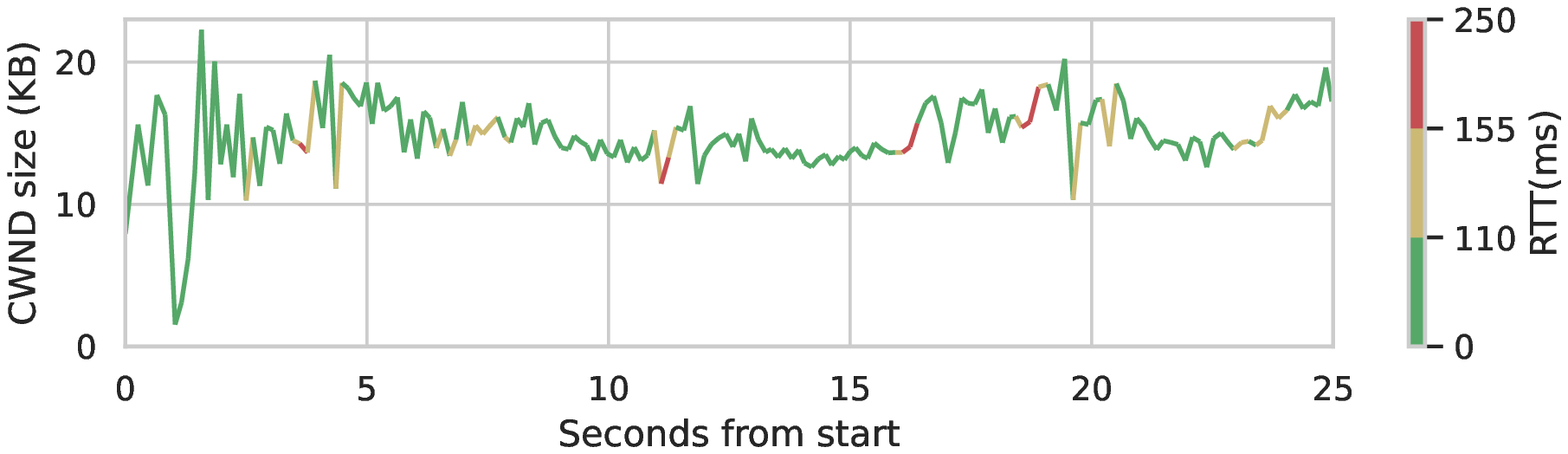}
\caption{The rolling averaged \gls{cwnd} size is plotted against the time from the beginning of the episode, while segment color represents the \gls{rtt} observed between two actions. This plot was obtained from the best run of the second MARLIN model, i.e. \ref{marlin_b}, which completed the transfer in 24.84s.}
\label{fig:marlin_cwnd}
\end{figure*}

\subsection{Future Directions}
The work described above is the first step of a multi-year research project. We have already identified future steps that follow naturally from our present work. While we proceed describing them sequentially, in reality these steps are much intertwined and we will likely addressed them in subsequent iterations.

We think that MARLIN would benefit from a more expressive reward function. We envision problem and reward formulations that truncate unpromising trajectories that have moved too distant from the optimal. We believe this could significantly speed up the solution convergence. Furthermore, we suspect that MARLIN's current state might present redundant information as well as features of low relevance to the problem. To address this, we plan to investigate smaller and more refined state representations, with the double goal of lowering complexity and improving convergence. We plan to train this new agent in diversified networking scenarios, which can capture different traffic patterns and network technologies, to assess the degree of generalizability. Finally, a thorough, automated hyperparameter tuning would further enhance MARLIN's performance and complete a first cycle of improvements. 

Further advancements will require an evolution in the agent's design congruent with specific problems that afflict \gls{cc}. The literature has shown that \gls{ml} models can accurately distinguish between packet loss attributable to congestion or channel errors. We plan to integrate a similar classifier within MARLIN and investigate the feasibility of an analogous approach to identify variations in end-to-end latency that are caused by changes in the path to destination. The fundamental building block of \gls{cc} algorithms is \textit{how and when} to change the \gls{cwnd} size. MARLIN currently actively controls the "how", leaving the "when" to a heuristic. We will investigate learning-based approaches to include such decision factor into MARLIN, with the goal of turning it into a more comprehensive and reactive system, able to make rational decisions at its own \textit{tempo}.

Up until now, we shaped the problem of \gls{cc} from the perspective of a single \gls{rl} agent. Nonetheless, the need for \gls{cc} algorithms in transport protocols originated from the lack of coordination in a multi-agent system, where single entities were acting in a completely self-centered manner. Therefore, we expect that the next step ahead in learning-based \gls{cc} will come from the application of advancing \gls{marl} algorithms that can optimize cooperative and/or competitive agent behavior.

%% file: tables/state.tex
\begin{tabular}{ c c l || c c }
\toprule
    & Feature & Description & & Statistic \\
\midrule
    1 & Current \gls{cwnd} &  Current \gls{cwnd} & 1 & Last \\
    2 & KBs Sent &  Amount of KB sent * & 2 & Mean \\
    3 & New KBs sent & Amount of KB acked * & 3 & STD \\
    4 & Acked KBs & Amount of KB acked *  & 4 & Min \\
    5 & Packets sent & Packets sent * & 5 & Max \\
    6 & Retransmissions & Number of packets retransmitted * & 6 & EMA \\
    7 & Instantaneous Throughput & Throughput * & 7 & Difference from Previous\ \\
    8 & Instantaneous Goodput & Goodput *  & & \\
    9 & Unacked KBs & Amount of KBs in flight & & \\
    10 & Last RTT & Last \gls{rtt} detected  *& & \\
    12 & Min RTT & Min \gls{rtt} * & & \\
    12 & Max RTT & Max \gls{rtt} *  & & \\
    13 & SRTT & Smoothed \gls{rtt} *  & & \\
    14 & VAR RTT &  \gls{rtt} variance * & & \\
       & &* During the last \gls{rtt} timeframe & & \\
\bottomrule
\end{tabular}

%% file: tables/hyperparameters.tex
\begin{tabular}{c c}
\toprule
    Hyperparameter & Value \\
\midrule
    Training steps & $1 \times 10^6$ \\
    History length & $10$ \\
    Training episode length & $200$ \\
    Learning rate & $3 \times 10^{-4}$ \\
    Buffer size & $5 \times 10^{5}$ \\
    Warm-up (learning starts) & $1 \times 10^{4}$ steps \\
    Batch size & 512 \\
    Tau & $5 \times 10^{-3}$ \\
    Gamma & $0.99$ \\
    Training Frequency & $1/episode$ \\
    Gradient Steps & $-1$ (same as episode length) \\
    Entropy regularization coefficient & "auto" (Learned) \\
    MLP policy hidden layers & $[400, 300]$\\
\bottomrule
\end{tabular}

%% file: sections/mockets_communication_protocol.tex
One of the main design choices we faced was the transport protocol we would use to train and evaluate MARLIN. The decision fell on a custom transport protocol because it is much simpler to integrate with MARLIN than \gls{tcp}. Additionally, a custom protocol enables greater flexibility in terms of retrieving the information required to encode the agent state.

For this study, we integrated MARLIN with Mockets, a message-based communication middleware implemented on top of UDP, following a school of thought similar to the one that drove the design of the QUIC protocol \cite{rfc9000} (for additional details on Mockets, the reader can refer to \cite{mockets1}). Mockets implements a very aggressive CC algorithm whose purpose is to fully utilize the available bandwidth in presence of variable communication latency and elevated packet loss. While it presents several advantages over TCP and other transport protocols in degraded environments \cite{mockets2}, Mockets' existing CC fails to share link capacity with other communication flows going through the same links and cannot adapt quickly to changes in the available bandwidth. Therefore, Mockets could benefit significantly from MARLIN, which could make it a viable choice also for traditional network scenarios.

        
To have an accurate representation of the state of the environment within MARLIN, we improved the \gls{rtt} estimation in Mockets. To do so, the sender keeps track of the transmission time of each packet. When an \gls{ack} arrives, the sender computes the \gls{rtt} by calculating the difference between the current time and the transmission time of the last packet received by the receiver before generating the \gls{ack} message and then subtracting the \gls{ack} processing time from it. To make this calculation possible, the receiver appends the identifier of the last received packet (a strictly increasing unsigned integer) and the processing time (in microseconds) to all \gls{ack} messages.


%% file: sections/experimental_results.tex

\subsection{Hardware Involved}
    The testbed described in Figure~\ref{fig:network} is implemented using \gls{cots} devices. The agent is trained on a workstation equipped with an Intel i7-8700 CPU, 32GB DDR4 RAM, an NVIDIA GeForce RTX 2060 GPU, and an Intel Wireless-AC 9560 \gls{nic}. The "Receiver Application" resides on a 4GB \gls{rpi}. The two \gls{ap} used are a TP-Link EAP245 and a Netgear WAC505, connected to two Netgear GS108E switches. A 2015 Apple MacBook Pro with macOS Catalina v10.15.6 generates background traffic, which is sent to the receiving application running on a Dell Latitude 7000 laptop (E7470). Both the desktop machine and E7470 run the Ubuntu 20.04.5 LTS OS. All \gls{nic}-related optimizations, e.g., TCP Segmentation Offload (TSO) or Large Receive Offload (LRO), are enabled, as per default, on all systems. A Ubiquiti Networks EdgeRouter X with EdgeOS routes packets between the two subnetworks.
    
\subsection{Training}
    The agent is trained for 1M steps on an infinite-horizon task with partial episodes lasting 200 steps. When the last step of a partial episode is encountered, a \textit{truncated} signal is emitted and \gls{peb} is performed, recalling the agent that additional steps and rewards would actually be available thereafter~\cite{time_limit}.
    
    Due to time constraints, the optimization steps involved (for actor, critic, target networks, and learned entropy coefficient) happen at the end of every partial episode. Taking training steps between two actions on a real network is a luxury we cannot afford, as we observed it would introduce delays in the order of ~800-1000ms. 
    Training at the end of an episode allows us to avoid such effect, limiting the overhead to the sole pipeline, which we report being around 25 to 30 ms. 
    
    At the beginning of training, in order to increase exploration, actions coming from a uniform random distribution are taken for 10K steps. Subsequently, \gls{sac} starts its normal exploration-exploitation strategy, led by its entropy coefficient.
    
    We trained three different models for MARLIN. For the first one, we used Eq~\ref{eq:reward_1} (\ref{marlin_a}). For the second model, we used the reward function in Eq.~\ref{eq:reward_2} (\ref{marlin_b}). In both cases, the background traffic pattern was fixed, with elephant flows following this order: $[100~\text{UDP}, 200~\text{TCP}, 100~\text{UDP}, 50~\text{UDP}]$. For the last model (\ref{marlin_c}), we changed the background traffic to use all permutations presented in Section~\ref{subsec:training_env_setup}.
    
    \begin{figure}[t]
    \centering
        \includegraphics[width=0.89\columnwidth]{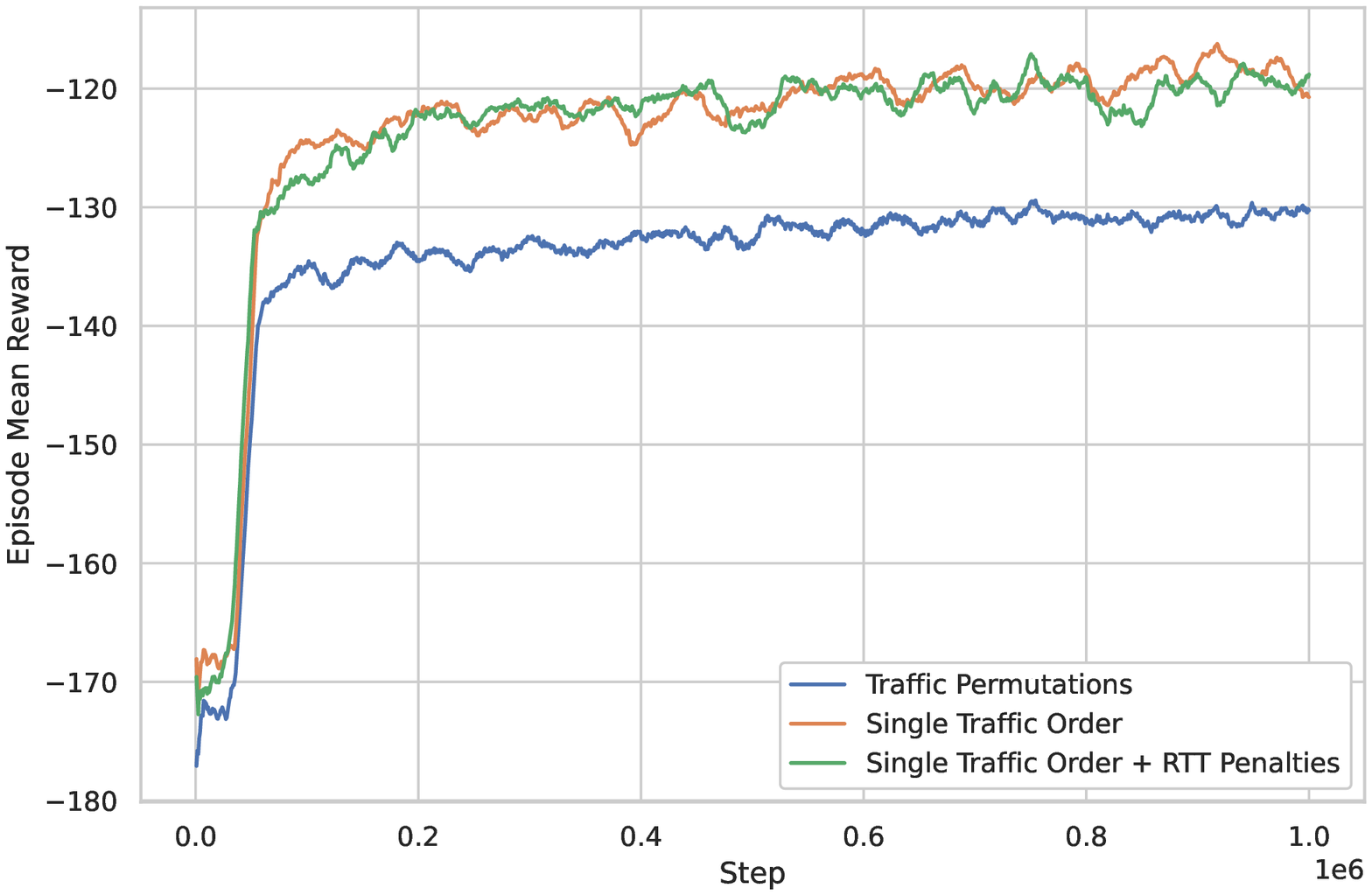}
        \caption{Mean training reward over the last 100 partial episodes on the three training runs.}
        \label{fig:reward}
    \end{figure}

    The mean training reward shows a promisingly increasing trend in each of the three setting, as shown in Figure \ref{fig:reward}. It is straightforward to interpret the performance showed during training by taking as a reference our ideal target, which, in this case, would result in -100 mean training reward.
    
\subsection{Results}
    Following training, we tested all the three MARLIN models by transferring a 3MB file on the same network with the background traffic pattern $[100~\text{UDP}, 200~\text{TCP}, 100~\text{UDP}, 50~\text{UDP}]$. It is important to note that file transfer is a significantly different problem from the one that the agent faces during training: here, the agent is evaluated on completing a task longer than the partial episodes seen during training, which are limited to 200 steps, and a faster completion coincides better results.
    
    We repeated the experiment 100 times for each of the three trained agents. Results are then compared to the performance of \gls{scp}, which in our system uses \gls{tcp} CUBIC as the transport protocol, on the same file transfer task.
    Figure~\ref{fig:trajectories} shows the optimal behaviour, computed from the a-priori knowledge of the background traffic, along with the batch of results obtained from the trained agents. The average, best, and worst performance of \gls{tcp} CUBIC are also shown under the same conditions after having repeated the transfer 100 times as well. Figure~\ref{fig:marlin_cwnd} presents the variation of the \gls{cwnd} size during MARLIN's best run across all the three experiments. The color of each segment represents the \gls{rtt} of the communication measured during a period of one \gls{srtt} immediately following the action that took the \gls{cwnd} size to the value represented by the rightmost end of the segment.
    
    Runs with significant performance degradation were aborted after 80 seconds, with \ref{marlin_a} reporting 4 aborted experiments, \ref{marlin_b} reporting 20, and \ref{marlin_c} reporting 2. 
    TCP CUBIC completes the file transfer in 45.03s in average, with its fastest transfer achieved in 33.97s, and a worst performance of 70.78s.
    Excluding aborted runs, MARLIN reached the target in 46.1s in average in \ref{marlin_a}, 50.31s in \ref{marlin_b}, and 46.9s in \ref{marlin_c}. All agents' best performance was faster than TCP CUBIC's best. In \ref{marlin_a}, 48\% of the experiments achieve a performance equal or better than the average accomplished by \gls{tcp} CUBIC, 31\% in \ref{marlin_b}, and 47\% in \ref{marlin_c}. The fastest transfer was completed by \ref{marlin_b} in 24.84s, 27\% faster than the fastest \gls{scp} transfer and 45\% faster than its average. For comparison, the available bandwidth on the link allows file transfers to be completed in 25s (see Figure~\ref{fig:trajectories}); faster transfers can still occur if the traffic injected by MARLIN causes other flows to slow down.
    
\subsection{Discussion}
    Despite the increased decision making complexity brought by continuous actions in a problem as intricate as \gls{cc} in a real network and a training budget of 1M steps, which is modest when compared to other \gls{rl}-based \gls{cc} agents, results are promising and already comparable to TCP CUBIC in our environment. We believe that part of it is due to the sample efficiency of \gls{sac}. Another reason can be found in the shaping of the \gls{cc} problem as an infinite-horizon task with strictly negative rewards, which enables the agent to exploit its acquired experience in tasks longer than the partial episodes seen during training and with different goals.
    
    Permuting the order of the background traffic patterns during training did not deteriorate the performance during evaluation. In fact, evaluation runs exhibited lower variance, a better fastest transfer, and similar average transfer time compared to the experiments shown in \ref{marlin_a}. Moreover, fewer experiments had to be aborted. These results suggest that MARLIN's robustness might be further improved by feeding data from diverse scenarios to the algorithm.

    Figure~\ref{fig:marlin_cwnd} shows the \gls{cwnd} size during one of the evaluation runs. Note that most of the time, following an increased \gls{rtt} reading, the agent has learned to respond by reducing the \gls{cwnd} size or decreasing its growth speed in the next step. This behavior is compatible with many \gls{rtt}-based \gls{cc} algorithms.
    
    Finally, although most trajectories obtained during evaluation express promising and valuable results (Figure~\ref{fig:trajectories}), they also present a significant degree of instability, which warrants additional research. This behavior is particularly evident in \ref{marlin_b}, where the model is trained using the function in Eq.~\ref{eq:reward_2}. This model has had several transfers significantly faster than the ones in \ref{marlin_a} and \ref{marlin_c}, as well as those obtained with \gls{scp}; nonetheless, the model has proved considerably slower in average.
    


%% file: sections/related_work.tex
Research efforts on the design and optimization of \gls{tcp} \gls{cc} have historically been very different from the approach discussed in this paper. Traditional \gls{cc} algorithms aim at achieving full bottleneck link utilization by applying diverse heuristic strategies that increase the \gls{cwnd} size until a congestion sign emerges, such as a packet loss, e.g., NewReno~\cite{rfc6582} and CUBIC~\cite{cubic} (the default CC algorithm in modern Linux Kernels and recent Windows operating systems), or an increase in latency, e.g., Vegas~\cite{Brakmo1994TCPVN} and, more recently, BBR~\cite{bbr}. These approaches have been shown to work well under specific network conditions, but underperform or experience other types of issues in other scenarios \cite{Bruhn2022, sosic2013, ha2021}.

Due to the exceptional degree of heterogeneity, complexity, and intrinsic dynamism of networking environments, and following the reinvigorated interest in \gls{ml} that captivated the scientific community in the last two decades, several recent efforts have focused on learning-based transport protocol optimization techniques. Some approaches focus on addressing very specific problem of transport protocols, such as accurately identifying congestion events, but do not aim at replacing \gls{cc} algorithms. For instance, researchers have successfully built \gls{ml}-based models that can distinguish between losses caused by congestion and losses due to channel errors in wireless networks \cite{ml-lda, loss_dtb} or losses caused by medium contention in optical burst switching networks \cite{nbbf-obs}. While we expect that our approach would benefit from such solutions, as discussed above, the final goal of MARLIN is to train an agent that can take on the tasks of \gls{cc} algorithms efficiently in a number of different network scenarios.

More interesting studies seek to apply \gls{ml} models and \gls{rl} agents directly to \gls{cc}, by learning policies that can optimally adjust the sending rate \cite{pcc}, the \gls{cwnd} \cite{pantheon}, or other parameters that tune the \gls{cc} algorithm \cite{tcp_remy}. Comprehensive surveys of applications of \gls{ml} to \gls{cc} are given in \cite{Wei2021, Jiang2021}. Common problems that emerged from the reviewed studies include parameter selection, computational complexity, high memory consumption, low training efficiency, and compatibility and fairness against existing \gls{cc} heuristics.

The possibility of training a \gls{rl} agent in a completely autonomous manner makes the case for applying \gls{rl} to \gls{cc} optimization very compelling. The blocking nature of widely used \gls{rl} libraries such as OpenAI Gym, designed to solve \gls{rl} problems such as the ATARI games~\cite{openai-gym}, which can be blocked while waiting for the next action, constitutes one of the main challenges. Researchers have circumvented this obstacle by training agents in simulated environments and achieved very promising results, such as with DRL-CC~\cite{drl-cc} and Aurora~\cite{aurora}. However, reproducing those results in real-world networks has proved to be a challenge. Solutions that rely on blocking while the agent computes the next action introduce significant delays, which are detrimental to the overall communication performance, especially in fast networks. In contrast, we designed MARLIN to integrate asynchronously with Mockets, which avoids blocking the transport protocol waiting for the agent to take the next action.


Other researchers propose the usage of non-blocking agents to optimize \gls{cc}. MVFST-RL~\cite{Sivakumar2019} proposed a non-blocking agent based on IMPALA~\cite{impala}, a C++ implementation of the QUIC transport protocol, and Pantheon~\cite{pantheon} for network emulation. Communication between the agent and the system work in a similar fashion to Park~\cite{NEURIPS2019_f69e505b}, a platform for experimenting with \gls{rl} agents on computer system problems based on \glspl{rpc}. MARLIN follows a similar philosophy to avoid the drawbacks of blocking systems. However, to the best of our knowledge, MARLIN, is the only work that uses strictly negative rewards and investigates the use of an off-policy and entropy-regularized \gls{rl} algorithm, such as \gls{sac}, with a continuous action space, trained on a real network in which real background traffic flows compete for bandwidth access. Additionally, we trained MARLIN on a infinite-horizon setting and evaluated the model on a common real-world problem such as transferring a file over a shared link. 

%% file: sections/conclusion.tex
This work has shown how effective policies can be obtained by training a \gls{rl} agent based on a off-policy, entropy-regularized algorithm such as \gls{sac}. MARLIN shapes the \gls{cc} problem as a strictly negative rewarded task actuating on continuous-actions in a real network with competing dynamic background traffic. 

We have also presented future research directions that we plan to pursue. These include training in more heterogeneous environments, exploring \gls{marl} settings, investigating more expressive reward functions, and designing an agent able to autonomously decide when to take the next action.
